\documentclass[11pt,a4paper]{article}
\usepackage{authblk}
\usepackage[nohyperref]{acl2017}
\usepackage{times}
\usepackage{latexsym}
\usepackage{url}
\usepackage{multirow}
\usepackage{rotating}
\usepackage{amsmath}
\usepackage{graphicx}
\usepackage{color}
\usepackage[footnotesize]{caption}
\usepackage{placeins}
\usepackage{bbm}
\usepackage{sidecap}
\usepackage{wrapfig}
\usepackage[normalem]{ulem}

\aclfinalcopy 
\def\aclpaperid{3484}

\title{Automatic Compositor Attribution in the First Folio of Shakespeare}

\author[$*$]{Maria Ryskina}
\author[$\dagger$]{Hannah Alpert-Abrams}
\author[$\ddagger$]{Dan Garrette}
\author[$*$]{Taylor Berg-Kirkpatrick}
\affil[$*$]{ Language Technologies Institute, Carnegie Mellon University, {\tt \{mryskina,tberg\}@cs.cmu.edu}}
\affil[$\dagger$]{Comparative Literature Program, University of Texas at Austin, {\tt halperta@gmail.com}}
\affil[$\ddagger$]{ Google, {\tt dhgarrette@google.com}}

\begin{document}
\maketitle

\begin{abstract}
\textit{Compositor attribution}, the clustering of pages in a historical printed document by the individual who set the type, is a bibliographic task that relies on analysis of orthographic variation and inspection of visual details of the printed page. In this paper, we introduce a novel unsupervised model that jointly describes the textual and visual features needed to distinguish compositors. Applied to images of Shakespeare's \textit{First Folio}, our model predicts attributions that agree with the manual judgements of bibliographers with an accuracy of 87\%, even on text that is the output of OCR.

\end{abstract}

\section{Introduction}

Within literary studies, the field of bibliography has an unusually long tradition of quantitative analysis. One particularly relevant area is that of \textit{compositor attribution}---the clustering of pages in a historical printed document by the individual (the \textit{compositor}) who set the type. 
Like stylometry, a long-standing area of NLP that has largely focused on attributing the authorship of text \citep{holmes1994authorship,hope1994authorship,juola2006authorship,Koppel:2009:CMA:1484611.1484627,Jockers01062010}, the analysis of \textit{orthographic} patterns is fundamental to compositor attribution.
Additionally, compositor attribution often makes use of \textit{visual} features, such as whitespace layout, introducing new challenges.
These analyses have traditionally been done by hand, but efforts are painstaking due to the difficulty of manually recording these features.

In this paper, we present an unsupervised model specifically designed for compositor attribution that incorporates both textual and visual sources of evidence traditionally used by bibliographers \citep{hinman1963printing,taylor1981shrinking,blayney1991first}. 
Our model jointly describes the patterns of variation both in orthography and in the whitespace between glyphs, allowing us to cluster pages by discovering patterns of similarity and difference. 
When applied to digital scans of historical printed documents, our approach learns orthographic and whitespace preferences of individual compositors and predicts groupings of pages set by the same compositor.\footnote{The validity of compositor attribution has sparked an ongoing and heated debate among bibliographers \citep{mckenzie1969,mckenzie1984,rizvi2016}; while some reject parts or all of this approach, it continues to be cited in authoritative bibliographical texts \citep{gaskell2007,norton1996}. Without taking a position in this debate, we seek only to automate the methods that remain in use by particular bibliographers \citep{norton1996,burrows2013}.}
This is, to our knowledge, the first attempt to perform compositor  attribution automatically. Prior work has proposed automatic approaches to authorship attribution---which is typically viewed as the supervised problem of identifying a particular author given samples of their writing. In contrast, compositor attribution lacks supervision because compositors are unknown and, in addition, focuses on different linguistic patterns. We explain spellings of words conditioned on word choice, not the word choice itself.

\begin{figure}[!t]
\vspace{-0.4cm}
\begin{center}
\includegraphics[height=4.75cm]{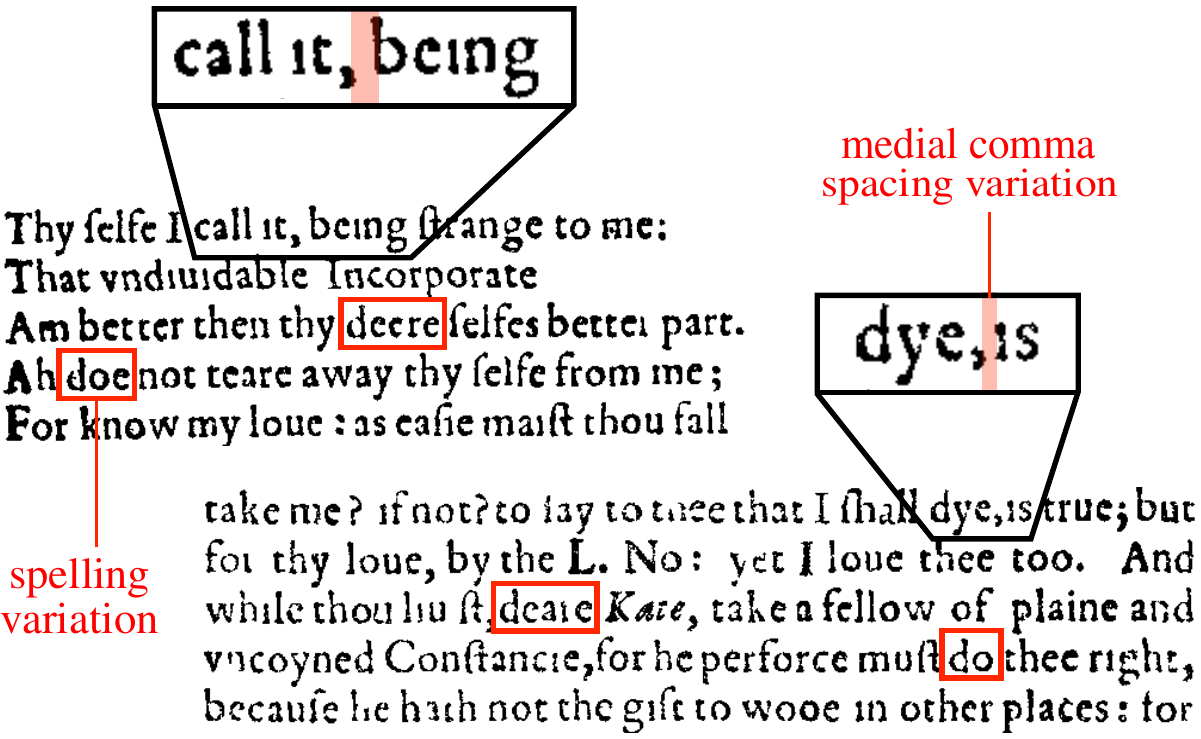}
\end{center}
\vspace{-0.2cm}
  \caption{\label{spelling-var} 
  The compositor of the left page tended to use the spellings \texttt{doe} and \texttt{deere}, while the compositor for the right page used spellings \texttt{do} and \texttt{deare}, indicating these pages were likely set by different people. The varying width of the medial comma whitespace also distinguishes the typesetters.}
  \vspace{-0.3cm}
\end{figure}

\begin{SCfigure*}[0.5][t]
\centering
\includegraphics[height=6.0cm]{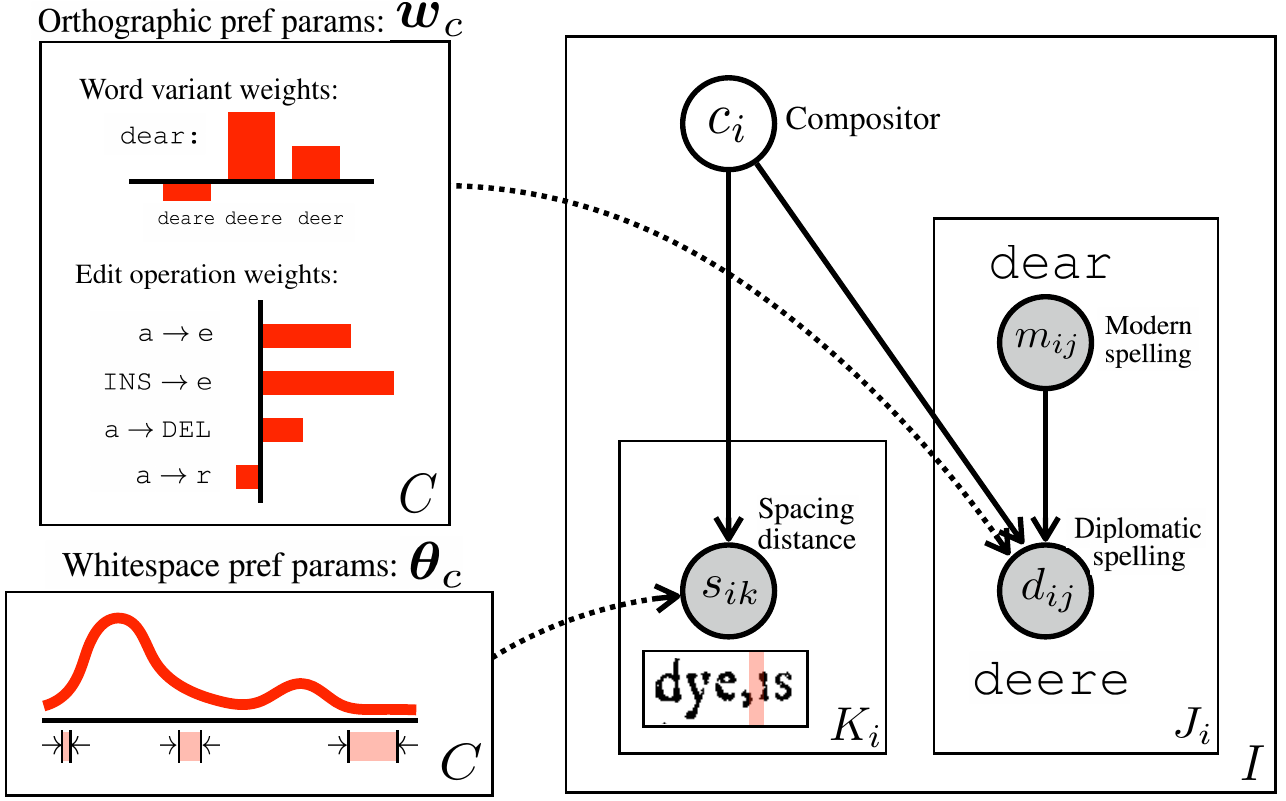}
\vspace{-0.2cm}
  \caption{\label{model-fig} In our model, a compositor $c_i$ is generated for page $i$ from a multinomial prior. Then, each diplomatic word, $d_{ij}$, is generated conditioned on $c_i$ and the corresponding modern word, $m_{ij}$, from a distribution parameterized by weight vector $\boldsymbol{w}_c$. Finally, each medial comma spacing width (measured in pixels), $s_{ik}$, is generated conditioned on $c_i$ from a distribution parameterized by $\boldsymbol{\theta}_c$.}
  \vspace{-0.1cm}
\end{SCfigure*}

To evaluate our approach, we fit our model to digital scans of Shakespeare's \textit{First Folio} (1623)---a document with well established manual judgements of compositor attribution. 
We find that even when relying on noisy OCR transcriptions of textual content, our model predicts compositor attributions that agree with manual annotations 87\% of the time, outperforming several simpler baselines. 
Our approach opens new possibilities for considering patterns across a larger vocabulary of words and at a higher visual resolution than has been possible historically.
Such a tool may enable scalable first-pass analysis in understudied domains as a complement to humanistic studies of composition.

\section{Background}
\label{sec:background}
In this paper we focus on modeling the same types of observations made by scholars and demonstrate agreement with authoritative attributions. 
We use compositor studies of Shakespeare's \textit{First Folio} to inform our approach, drawing on the methods proposed by \citet{hinman1963printing}, \citet{howard1973compositors}, and  \citet{taylor1981shrinking}. Hinman's landmark 1963 study clustered the pages of the \textit{First Folio} according to five different compositors based on variations in spelling among three common words. Figure~\ref{spelling-var}, for example, shows portions of two pages of the \textit{First Folio} with different spelling variants for the words \texttt{dear} and \texttt{do}: one compositor used \texttt{deere} and \texttt{doe}, while the other used \texttt{deare} and \texttt{do}.
Hinman relied on the assumption that each compositor was consistent in their preferences for the sake of convenience in the typesetting process~\cite{blayney1991first}.
Subsequent studies looked at larger sets of words and more general orthographic preferences (e.g. the preference to terminate words with \texttt{-ie} instead of \texttt{-y}), leading to modifications of Hinman's original analysis \citep{howard1973compositors, taylor1981shrinking}. 
In this paper we propose a probabilistic model designed to capture both word-specific preferences and general orthographic patterns. To separate the effect of the compositor from the choices made by the author or editor, we condition on a modernized (collated) version of Shakespeare's text as was done by scholars.

Visual features, including typeface usage and whitespace layout, also inform compositor attribution. For example, the highlighted spacing in Figure~\ref{spelling-var} shows different choices after medial commas (commas that occur before the end of the line). 
Bibliographers produced new hypotheses about how many compositors were involved in production based on the analysis of the use of spaces before and after punctuation~\citep{howard1973compositors, taylor1981shrinking}. We additionally incorporate this source of evidence into to our automatic approach by modeling pixel-level whitespace distances.

Bibliographers also use contextual information to inform their analyses, including copy text orthography, printing house records, collation, type case usage, and the use of type with cast-on spaces. In our model, we restrict our analysis to only those features that can be derived from the OCR output and simple visual analysis.

\vspace{-0.1cm}
\section{Model \label{model-sec}}

Our computational approach to compositor attribution operates on the sources of evidence that have been considered by bibliographers. In particular, we focus on jointly modeling patterns of orthographic variation and spacing preferences across pages of a document, treating compositor assignments as latent variables in a generative model. We assume access to a diplomatic transcription of the document (a transcription faithful to the original orthography), which we automatically align with a modernized version.\footnote{Modern editions are common for many books that are of interest to bibliographers, though future work could consider how to cope with their absence.} We experiment with both manually and automatically (OCR) produced transcriptions, and assume access to pixel-level spacing information on each page, which can be extracted using OCR as described in Section~\ref{experiments}.  

Figure~\ref{model-fig} shows the generative process. In our model, each of $I$ total pages is generated independently. The compositor assignment for the $i$th page is represented by the variable $c_i \in \{1, \ldots, C\}$ and is generated from a multinomial prior. For page $i$, each diplomatic word, $d_{ij}$, is generated conditioned on the corresponding modern word, $m_{ij}$, and the compositor who set the page, $c_i$. Finally, the model produces the pixel width of the space after each medial comma, $s_{ik}$, again conditioned on the compositor, $c_i$. The joint distribution for page $i$, conditioned on modern text, takes the following form:
\begin{align*}
P(& \{d_{ij}\}, \{s_{ik}\}, c_i | \{m_{ij}\} ) = \\
& P(c_i) \hspace{2.9cm} \text{[Prior on compositors]}\\
\cdot & \prod_{j=1}^{J_i} P(d_{ij} | m_{ij}, c_i; \boldsymbol{w}_{c_i}) \hspace{0.27cm} \text{[Orthographic model]}\\
\cdot & \prod_{k=1}^{K_i} P(s_{ik} | c_i; \boldsymbol{\theta}_{c_i}) \hspace{1.35cm} \text{[Whitespace model]}
\end{align*}

\subsection{\label{sec:spelling+ortho} Orthographic Preference Model}

We choose the parameterization of the distribution of diplomatic words in order to capture two types of spelling preference: (1) general preferences for certain character groups (such as \texttt{-ie}) and (2) preferences that only pertain to a particular word and do not indicate a larger pattern. 
Since it is unknown which of the two behaviors is dominant, we let the model describe both and learn to separate their effects.
Using a log-linear parameterization,
\[
P(d | m, c; \boldsymbol{w}) \propto \exp (\boldsymbol{w}_c^\top \mathbf{f}(m, d))
\]
we introduce features to capture both effects. Here, $\mathbf{f}(m, d)$ is a feature function defined on modern word $m$ paired with diplomatic word $d$, while $\boldsymbol{w}_c$ is a weight vector corresponding to compositor $c$.

To capture word-specific preferences we add an indicator feature for each pair of modern word $m$ and diplomatic spelling $d$. We refer to these as \textsc{word} features below. To capture general orthographic preferences we introduce an additional set of features based on the edit operations involved in the computation of Levenshtein distance between $m$ and $d$. In particular, each operation is added as a separate feature, both with and without local context (previous or next character of the modern word). We refer to this group as \textsc{edit} features. The weight vector for each compositor represents their unique biases, as shown in the depiction of these parameters in Figure~\ref{model-fig}.

\hspace{-0.2cm}\begin{SCtable*}[1][t]
\centering
\caption{\label{tab:results} The experimental results for all setups of the model. In the experiments with \textsc{basic} model, we compare the short \textsc{hinman} word list with the automatically filtered \textsc{auto} word list. We show results for several variants of our full model, labeled as \textsc{feat}, both with and without spacing generation. A random baseline is included for comparison. \vspace{-0.2cm}}
\scalebox{0.78}{
\begin{tabular}{|p{0.7cm} p{2.7cm}||cc|cc||cc|cc|}
     \hline
     \multicolumn{2}{|c||}{\multirow{ 3}{*}{Model Setup}} & \multicolumn{4}{c||}{Bodleian Transcription} & \multicolumn{4}{c|}{Ocular OCR Transcription} \\
     \cline{3-10}
     & & \multicolumn{2}{c|}{Hinman Attr} & \multicolumn{2}{c||}{Blayney Attr} & \multicolumn{2}{c|}{Hinman Attr} & \multicolumn{2}{c|}{Blayney Attr} \\
      & & 1-to-1 & M-to-1 & 1-to-1 & M-to-1 & 1-to-1 & M-to-1 & 1-to-1 & M-to-1 \\
     \hline
     \hline
     \textsc{random} & & 22.5 & 49.6 & 16.7 & 49.6 & 22.5 & 49.6 & 16.7 & 49.6 \\
     \hline
     \hline
     \textsc{basic} & w/ \textsc{hinman} & 67.9 & 71.8 & 60.4 & 67.3 & 66.6 & 70.5  & 47.1 & 63.8 \\
     \hline
     & w/ \textsc{auto} & 64.3 & 81.0 & 58.8 & 81.3 & 64.9 & 81.1 & 53.7 & 80.7 \\
     \hline
     \hline
     \textsc{feat} & w/ \textsc{edit} & 75.3 & 79.1 & 77.1 & 83.1 & 76.8 & 77.4 & 76.1 & 76.0 \\
     \hline
      & w/ \textsc{edit} + \textsc{word} & 81.1 & 81.1 & 80.7 & 80.6 & 75.1 & 75.0 & 74.4 & 74.4 \\
     \hline
      & w/ \textsc{edit} + \textsc{space} & \bf 87.6 & \bf 87.5 & \bf 87.3 & \bf 87.2 & \bf 86.7 & \bf 86.6 & \bf 85.9 & \bf 85.8 \\
     \hline
      & w/ \textsc{all} & 83.8 & 83.7 & 83.5 & 83.4 & 82.5 & 82.4 & 82.4 & 82.2 \\
     \hline
\end{tabular}}
\end{SCtable*}

\vspace{-0.5cm}
\subsection{Whitespace Preference Model}
\label{sec:spacing}

Manual analysis of spacing has revealed differences across pages. In particular, the choice of spaced or non-spaced punctuation marks is hypothesized by biobliographers to be indicative of compositor preference and specific typecase. We add whitespace distance to our model to capture those observations. While bibliographers only made a coarse distinction between spaced or non-spaced commas, in our model we generate medial comma spacing widths, $s_{ik}$, that are measured in pixels to enable finer-grained analysis. We use a simple multinomial parameterization where each pixel width is treated as a separate outcome up to some maximum allowable width:
\[
s_{ik} | c_i \sim Mult(\boldsymbol{\theta_{c_i}})
\]
Here, $\boldsymbol{\theta}_c$ represents the vector of multinomial spacing parameters corresponding to compositor $c$.
We choose this parameterization because it can capture non-unimodal whitespace preference distributions,
as depicted in Figure~\ref{model-fig}, and it makes learning simple.

\subsection{Learning and Inference}

Modern and diplomatic words and spacing variables are observed, while compositor assignments are latent. In order to fit the model to an input document we estimate the orthographic preference parameters, $\boldsymbol{w}_c$, and spacing preference parameters, $\boldsymbol{\theta}_c$, for each compositor using EM. The E-step is accomplished via a tractable sum over compositor assignments, while the M-step for $\boldsymbol{w}_c$ is accomplished via gradient ascent \cite{painless_2010}. The M-step for spacing parameters, $\boldsymbol{\theta}_c$, uses the standard multinomial update. Predicting compositor groups is accomplished via an independent argmax over each $c_i$. In all experiments we run 75 iterations of EM with 100 random restarts, choosing the learned parameters corresponding the best model likelihood.

\section{Experiments \label{experiments}}

\noindent {\bf Data:} To evaluate our model when it has access to perfectly transcribed historical text, we use the Bodleian diplomatic transcription of the \textit{First Folio}.\footnote{\url{http://firstfolio.bodleian.ox.ac.uk/}} To test whether our approach can also work with untranscribed books, we ran the Ocular OCR system~\citep{bergkirkpatrick-durrett-klein:2013:ACL2013} on the Bodleian facsimile images to create an automatic diplomatic transcription. In both cases, we used Ocular's estimates of glyph bounding boxes on the complete \textit{First Folio} images to extract spacing information. The modern text was taken from MIT Complete Works of Shakespeare\footnote{\url{http://shakespeare.mit.edu/}} and was aligned with diplomatic transcriptions by running a word-level edit distance calculation. The extracted substitutions form the model's observed modern-diplomatic word pairs.
\\[0.15cm]
\noindent {\bf Evaluation:} To compare the recovered attribution  with those proposed by bibliographers, we evaluate against an authoritative attribution compiled by Peter Blayney~\shortcite{norton1996} which includes the work of various scholars~\citep{hinman1963printing, howard1973compositors, howardhill1976,  howard1980new, taylor1981shrinking, o1975compositors, werstine1982cases}. We also evaluate our system against an earlier, highly influential model proposed by \citet{hinman1963printing}, which we approximate by reverting certain compositor divisions in Blayney's attribution. Hinman's attribution posited five compositors, while Blayney's posited eight. In experiments, we set the model's maximum number of compositors to \mbox{$C=5$} when evaluating on Hinman's attribution, and use \mbox{$C=8$} with Blayney's.
We compute the one-to-one and many-to-one accuracy, mapping the recovered page groups to the gold compositors to maximize accuracy, as is standard for many unsupervised clustering tasks, e.g. POS induction (see \citet{christodoulopoulos:emnlp2010}).
\\[0.15cm]
\noindent {\bf\textsc{basic} model variant:} We evaluate a simple baseline model that uses a multinomial parameterization for generating diplomatic words and does not incorporate spacing information. We use two different options for selection of spelling variants to be considered by the model. First, we consider only the three words selected by Hinman: \texttt{do}, \texttt{go} and \texttt{here} (referred to as \textsc{hinman}). Second, we use a larger, automatically selected, word list (referred to as \textsc{auto}). Here, we select all modern words with frequency greater than 70 that are not names and that exhibit sufficient variance in diplomatic spellings (most common diplomatic spelling occurs in less than 80\% of aligned tokens). For our full model, described in the next section, we always use the larger \textsc{auto} word list. 
\\[0.15cm]
\noindent {\bf\textsc{feat} model variant:} We run experiments with several variants of our full model, described in Section~\ref{model-sec} (referred to as \textsc{feat} since they use a feature-based parameterization of diplomatic word generation.) We try ablations of \textsc{word} and \textsc{edit} features, as well as model variants with and without the spacing generation component (referred to as \textsc{space}.) We refer to the  full model that includes both types of features and spacing generation as \textsc{all}.

\begin{SCfigure*}[30][!t]
\includegraphics[height=3.4cm]{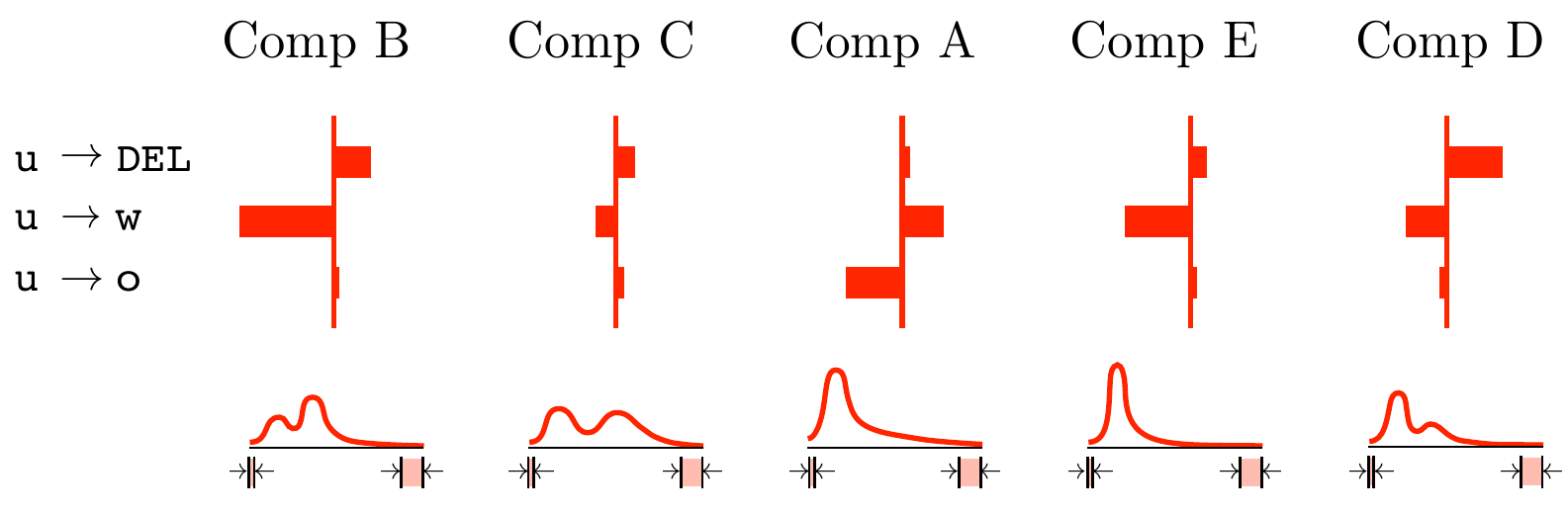}
 \caption{\label{results} 
Learned behaviors of the Folio compositors. Our model only detected the presence of five compositors (ranked according to number of pages the compositor set in our model's prediction).  Compositor D's habit of omitting~\texttt{u} (\texttt{yong} vs. \texttt{young}) and compositor C's usage of spaced medial commas were also noticed in~\citet{taylor1981shrinking}.} 
\end{SCfigure*}

\section{Results} 
\label{sec:results}

Our experimental results are presented in Table~\ref{tab:results}. The \textsc{basic} variant, modeled after Hinman's original procedure, substantially outperforms the random baseline, with the \textsc{hinman} word list outperforming the larger \textsc{auto} word list. However, use of the larger word list with feature-based models yields large gains in all scenarios, including evaluation on Hinman's original attributions and while using OCR diplomatic transcriptions. The best-performing model for both manually transcribed and OCR text uses \textsc{edit} features in conjunction with spacing generation and achieves an accuracy of up to 87\%. Including \textsc{word} features on top of this leads to slightly reduced performance, perhaps as a result of the substantially increased number of free parameters. In the OCR scenario, the addition of \textsc{word} features on top of \textsc{edit} decreases accuracy, unlike the same experiment with the manual transcription. This is possibly a result of the reduced reliability of full word forms due to mistakes in OCR.

Particularly interesting is the result that spacing, rarely a factor considered in NLP models, improves the accuracy significantly for our system when compared with \textsc{edit} features alone. Because pixel-level visual information and arbitrary orthographic patterns are also the most difficult features to measure manually, our results give strong evidence to the assertion that NLP-style models can aid bibliographers. 

\section{Discussion}

The results on OCR (character error rate for most plays $\approx 10-15\%$) transcripts are only marginally worse than those on manual transcripts, which shows that our approach can be generalized for the common case where manual diplomatic transcriptions are not available. For our experiments, we also chose a common modern edition of Shakespeare instead of more carefully produced modernized transcription of the facsimile---our goal being to again show that this approach can be generalized, perhaps to documents where careful modernizations of the facsimile are not available. Together, these results suggest that our model may be sufficiently robust to aid bibliographers in their analysis of less studied texts.

Figure~\ref{results} shows an example of the feature weights and spacing parameters learned by the \textsc{feat} w/ \textsc{all} model. Our statistical approach is able to successfully
explain some of the observations scholars made. For example,~\citet{taylor1981shrinking} notices that compositors C and D prefer to omit \texttt{u} in \texttt{young} but A does not. Our model reflects this by giving \texttt{u $\rightarrow$ DEL} high weight for D and low weight for A. However, the weight of a single feature is difficult to interpret in isolation. This might be the reason why our model only moderately agrees in case of compositor C. Another example can be seen in spacing patterns: according to~\citet{taylor1981shrinking}, compositor C uses spaced medial commas unlike A and D. Our model learns the same behavior.

\section{Conclusion}

Our primary goal is to scale the methods of compositor attribution, including both textual and visual modes of evidence, for use across books and corpora. 
By using principled statistical techniques and considering evidence at a larger scale, we offer a more robust approach to compositor identification than has previously been possible. The fact that our system works well on OCR texts means that we are not restricted to only those documents for which we have manually produced transcriptions, opening up the possibility for bibliographic study on a much larger class of texts. Though we are unable to incorporate the kinds of world knowledge used by bibliographers, our ability to include more information and more fine-grained information allows us to recreate their results. Having validated these techniques on the \textit{First Folio}, where historical claims are well established, we hope future work can extend these methods and their application.

\section*{Acknowledgements}
 We thank the three anonymous reviewers for their valuable feedback. This project is funded in part by the NSF under grant 1618044.

\bibliographystyle{acl_natbib}

\bibliography{refs}

\end{document}